\documentclass[10pt,twocolumn,letterpaper]{article}

\usepackage{cvpr}
\usepackage{times}
\usepackage{epsfig}
\usepackage{graphicx}
\usepackage{amsmath}
\usepackage{amssymb}
\usepackage[ruled]{algorithm2e}
\usepackage{enumitem}
\usepackage{amsmath}
\usepackage{enumitem}
\setitemize{noitemsep,topsep=0pt,parsep=0pt,partopsep=0pt}

\usepackage[pagebackref=true,breaklinks=true,letterpaper=true,colorlinks,bookmarks=false]{hyperref}

\cvprfinalcopy 

\def\cvprPaperID{6250} 

\ifcvprfinal\fi
\begin{document}
\setlength{\abovedisplayskip}{0pt}
\setlength{\belowdisplayskip}{0pt}
\setlength{\abovedisplayshortskip}{0pt}
\setlength{\belowdisplayshortskip}{0pt}

\newcommand\blfootnote[1]{%
  \begingroup
  \renewcommand\thefootnote{}\footnote{#1}%
  \addtocounter{footnote}{-1}%
  \endgroup
}

\title{Temporal Transformer Networks: \\Joint Learning of Invariant and Discriminative Time Warping}

\author{Suhas Lohit \qquad Qiao Wang \qquad Pavan Turaga\\
[4pt]
Geometric Media Lab, Arizona State University\\
[2pt]
{\tt \small \{slohit, qiao.wang, pturaga\}@asu.edu}
\\[-3pt]
}

\maketitle
\thispagestyle{empty}

\begin{abstract}
   Many time-series classification problems involve developing metrics that are invariant to temporal misalignment. In human activity analysis, temporal misalignment arises due to various reasons including differing initial phase, sensor sampling rates, and elastic time-warps due to subject-specific biomechanics. Past work in this area has only looked at reducing intra-class variability by elastic temporal alignment. In this paper, we propose a hybrid model-based and data-driven approach to learn warping functions that not just reduce intra-class variability, but also  increase inter-class separation. We call this a temporal transformer network (TTN). TTN is an interpretable differentiable module, which can be easily integrated at the front end of a classification network. The module is capable of reducing intra-class variance by generating input-dependent warping functions which lead to rate-robust representations. At the same time, it increases inter-class variance by learning warping functions that are more discriminative. We show improvements over strong baselines in 3D action recognition on challenging datasets using the proposed framework. The improvements are especially pronounced when training sets are smaller.
\end{abstract}

\section{Introduction}
\label{sec:intro}

Guaranteed invariances of machine learning algorithms to nuisance parameters is an important design consideration in critical applications. Classically, invariances can only be guaranteed under a model-based approach. Learned representations however have not been able to guarantee invariances, except by empirical tests~\cite{goodfellow2009measuring}. Learning invariant representations that build on analytical models of phenomena may hold the cue to bridge this gap, and can also help lend explainability to the model. \blfootnote{* Qiao Wang is now at SRI International.}

However, deep learning presents a challenge for learning explainable invariants, primarily due to incompatibility between the mathematical approaches that underly invariant design, and the architectures prevalent in deep learning. There have been recent attempts at leveraging model-based and data-driven approaches to learn invariant representations across spatial transforms~\cite{khasanova2017graph,kanazawa2014locally,xu2014scale}, illumination~\cite{tang2012deep,lohit2017learning}, and viewpoints~\cite{liu2017enhanced}. On the other hand, learning invariant/robust representations to temporal rate-variations has received significantly less attention. If tackled well, many applications of human activity modeling will benefit, including more robust recognition algorithms for human-robot interaction, richer synthesis of human motion for computer-generated imagery, and health applications. 

\textbf{Hybrid model- and data-based approach:} In this paper, our chosen application is activity classification from RGBD devices, where skeleton data are available. Activities such as walking can be performed at different rates by different subjects owing to physiological and biomechanical factors. We would like to design representations that provide robust classification against such nuisance factors. To do this, we adopt a model-based approach, and constrain certain layers in deep network using the model. The model for temporal variability is adopted from past work in elastic temporal alignment which considers temporal variability as a result of a temporal diffeomorphism acting on a given time-series~\cite{srivastava2016functional}. The space of such diffeomorphisms has a group structure, and can be converted to simpler geometric constraints by exploiting contemporary square-root forms to represent the diffeomorphic maps~\cite{srivastava2007riemannian}.

\begin{figure*}
    \centering
    \includegraphics[trim={1cm 1.5cm 1cm 1cm},clip,width=0.88\textwidth]{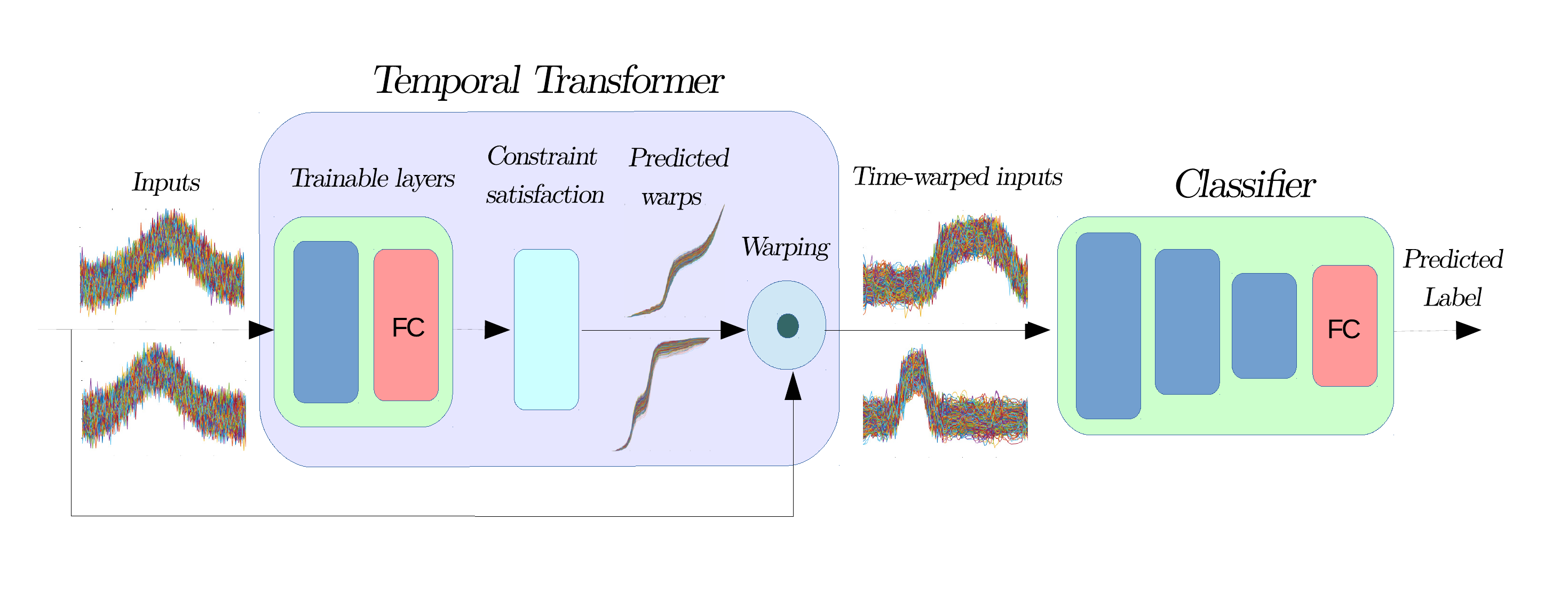}
    \caption{The Temporal Transformer Network (TTN) is a trainable module that is easily added at the beginning of a time-series classifier. Its function is to warp the input sequences so as to maximize the classification performance, as shown in the figure for two classes of waveforms which become more discriminative after passing through the TTN. The sub-modules of the TTN are explained in Section \ref{sec:temporal_transformer}.}
    \label{fig:ttn_overview}
    \vspace{-0.1in}
\end{figure*}

\textbf{Compatibility with deep architectures:} We design a novel module, which we refer to as a Temporal Transformer Network (TTN). The hallmark of this module is that it can be easily integrated into existing time-series classifiers such as Temporal Convolution Networks (TCNs)~\cite{kim2017interpretable} and Long Short-Term Memory (LSTM) networks~\cite{hochreiter1997long}. TTN is explainable in the sense that it is designed so as to interact with the classification network in a predefined, predictable and visualizable manner. TTN is a trainable network added at the beginning of the classifier and operates on the input sequence by performing selective temporal warping of the input sequences. As such, it has the ability to factor out rate variations, if present in the data, as well as increase the inter-class separation by learning to align sequences in dissimilar classes away from each other.

\textbf{Application impact:} Recognition of human activities from motion capture (mocap) systems such as OptiTrack or depth sensors like Microsoft Kinect and Intel RealSense has been gathering a great deal of interest in the recent past. The cost of these sensors is ever-reducing and the increasing effectiveness of pose estimation algorithms~(e.g.~\cite{toshev2014deeppose}) makes 3D skeletons an important sensing modality for action recognition. As the problem of action recognition presents a large amount of variability both inter-class as well as intra-class, we choose it as the focus of this paper.

\vspace{-0.15in}
\paragraph{Contributions}
\begin{itemize}[leftmargin=*]
\setitemize{noitemsep,topsep=0pt,parsep=0pt,partopsep=0pt}
    \item We propose the Temporal Transformer Network (TTN), which performs joint representation learning as well as class-aware discriminative alignment for time-series classification including action trajectories.
    \item We design the TTN to generate highly expressive non-parametric, order-preserving diffeomorphisms.
    \item The TTN exploits the non-uniqueness of the optimal alignment (between equivalence classes) to generate discriminative warps for improved classification.
     \item The proposed TTN can be easily integrated into existing time-series classification architectures, with just a single line of code to employ the warping module. 
\end{itemize}
We validate our contributions by demonstrating improved performance on small and large, and real and synthetic datasets, for action recognition from 3D pose obtained from two different modalities -- Kinect and mocap. The combined architecture of the TTN and the classifier consistently yields improved classification performance compared to several baseline classifiers.

\section{Related work}
\label{sec:related_work}

\textbf{Deep learning of invariant representations:}
One of the main inspirations for this work is the paper by Jaderberg et al.~\cite{jaderberg2015spatial} on Spatial Transformer Networks (STNs) where a smaller network first predicts a geometric transform of the input grid parameterized by affine transforms or thin plate splines. The transformation is then applied to the input before feeding it to the classification network. A recent work is that of Detlefsen et al.~\cite{skafte2018deep} who improve the performance of spatial transformers by replacing affine transforms and thin plate splines with a richer class of parameterized diffeomorphic transforms called continuous piecewise-affine transforms, but at the expense of complex implementation and considerably longer training times. Both these works are aimed at building invariances to spatial geometric transforms of images. Capsule networks by Sabour et al.~\cite{sabour2017dynamic} expand the expressive capacity of CNNs by allowing them to learn explicit spatial relationships. An interesting recent work by Tallec and Ollivier~\cite{tallec2018can} show that LSTM networks have the capability to learn to warp input sequences. Our experiments show that by integrating LSTMs with the module designed in this paper, the performance can be further increased, as the proposed framework can also lead to more discriminative representations.

In this paper, we design a module to predict warping functions in the temporal domain which when applied to the input sequences lead to higher classification performance. This requires the predicted warping functions to satisfy the order-preserving property. Moreover, in our case, the predicted warping functions are non-parameterized, can span the entire space of rate-modifying transforms and are much more expressive than earlier related works. The warping is also elastic, as opposed to rigid deformations which is the case with STNs. We note that these are important design requirements in the case of temporally varying signals, that are different from transforms in the 2D spatial domain.

\textbf{Alignment of time-series data:} The most commonly used method to align time-series data is Dynamic Time Warping (DTW)~\cite{bellman1958adaptive, sakoe1978dynamic}. DTW tries to minimize the $\mathbb{L}^2$ distance between two time series after a time warping is applied to one of them, and is agnostic of class information. To address some of DTW's shortcomings, new methods have been proposed recently, including the elastic functional data and shape analyses~\cite{srivastava2011shape, srivastava2016functional} which defined proper metrics that are invariant to time warping, and soft-DTW which is a differentiable loss function that may be integrated into neural networks~\cite{cuturi2017soft}. There are also several works on Canonical Time Warping (CTW) for multimodal data where the time series from different streams are  projected to a common space (similar to canonical correlation analysis) before aligning them~\cite{zhou2016generalized}. Deep learning versions of CTW have also been proposed recently~\cite{trigeorgis2016deep, trigeorgis2018deep, jia2017sparse}.

One of the major differences between our proposed method and the aforementioned optimization-based time warping methods is that our approach performs discriminative warping based on class information and does not need signal templates. This is further discussed in Section~\ref{sec:temporal_transformer}. During the preparation of this document, we came across the paper by Oh et al.~\cite{oh2018learning} who propose a similar architecture for clinical time series data classification but restricted to the space of linear time scaling and offsets.

\textbf{3D action recognition using deep learning:} As sensing systems like Microsoft Kinect, Intel RealSense and camera-based mocap systems are getting more effective at acquiring depth and at human pose estimation with centimeter- to millimeter-level accuracy, research and commercial interest in employing 3D pose data for action recognition has understandably increased. Recent experiments suggest that for small datasets, recognition accuracies are better with 3D pose information compared to video frames~\cite{FirstPersonAction_CVPR2018}. That simple landmark-based or skeleton-based action recognition can be effective is supported by evidence from works in psychology which show that humans are excellent at recognizing actions only from a few points on the human body~\cite{johansson1973visual}.

Recurrent neural architectures, especially Long Short-Term Memory (LSTM) networks have been used to perform 3D action recognition e.g.~\cite{du2015hierarchical,Shahroudy_2016_CVPR}. Song et al. propose including layers for spatial and temporal attention (STA-LSTM)~\cite{song2017end} which greatly improves the recognition performance. For majority of the experiments in this paper, we will use the Temporal Convolution Network (TCN) with residual connections~\cite{lea2016temporal} as they are effective, simple to build and faster to train compared to LSTM-based networks. Additionally, Kim and Reiter have shown excellent results on using TCNs for 3D action recognition~\cite{kim2017interpretable}. This network outperforms STA+LSTM~\cite{song2017end} for 3D action recognition. They further show that TCNs can learn both spatial and temporal attention without the need for special attention layers. Also, the network filter activations are interpretable by design because of the residual connections. We also note that the TCN architecture presented in~\cite{kim2017interpretable} incorporates pooling mechanisms inside the network. 

We note that more recently, newer architectures have proposed modifications to baseline architectures by using graph convolutions to better take into account the spatial structure of the joints in the human body~\cite{yan2018spatial}. However, it has a much higher computational load. Other representations include image-based ones~\cite{ke2017new, liu2017enhanced}, and fusing skeleton information with velocity information~\cite{li2017skeleton} etc. Our contributions in this paper are orthogonal to these works, and the main focus of the paper is to design a specialized module for learning rate-robust discriminative representations. As such, for our experiments, we choose two effective widely used simple-to-implement architectures as our baselines -- TCNs and LSTMs -- and demonstrate improvements in recognition performance over these frameworks.

\begin{figure}[b!]
    \centering
    \includegraphics[width=0.37\textwidth]{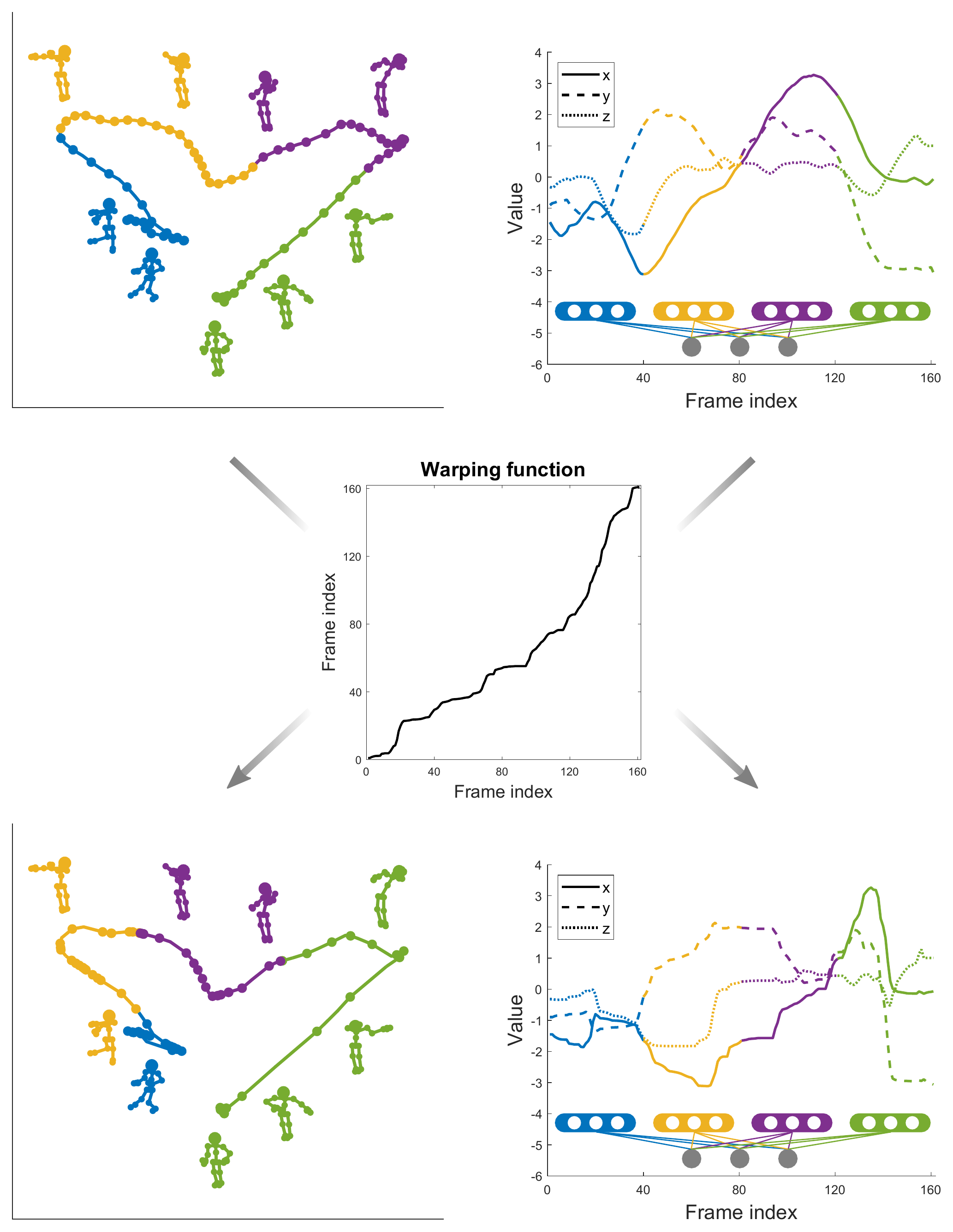}
    \caption{Top-left, Bottom-left: the trajectories and sampling points of the same action (``wearing jacket'') before and after a time warping (Center). The parameterized trajectories are visualized in $\mathbb{R}^3$ by using the sums of $x, y, z$ coordinates of all joints. Notice how the time series of $x, y, z$, which are the inputs of a neural network (Top-right and Bottom-right), are quite different despite the action being the same. Here the action is segmented and colored to highlight the rate variation.}
    \label{fig:curve_in_R_n}
    \vspace{-0.1in}
\end{figure}

\section{Diffeomorphic models for rate variation}
A continuous time-series can be represented as a single-parameter curve, which we denote by $\alpha(.)$, where $t \in [0,1]$ is the parameter. In our case, $t$ is time and we assume $\alpha(t) \in \mathbb{R}^N$. Another curve $\beta$ is a \textit{resampling} of $\alpha$ if $\beta = \alpha \circ \gamma$, where $\circ$ is a function composition, and $\gamma$ is the \textit{resampling/warping function}. We focus on a specific set $\Gamma$ of these warping functions (defined below), and two curves $\alpha_1, \alpha_2$ differing only by a change of rate of execution obey the equation $\alpha_1 = \alpha_2 \circ \gamma, \text{ for some } \gamma \in \Gamma$. Figure~\ref{fig:curve_in_R_n} shows an example of such time warping, illustrated in $\mathbb{R}^3$.    

Given a 1-differentiable function $\gamma$ defined on the domain $[0,T]$, for $\gamma$ to be an element of $\Gamma$, $\gamma$ needs to satisfy the following conditions:
\begin{align} 
\gamma(0) &= 0 , \gamma(1) = 1, 
\textrm{ and } \gamma(t_1) < \gamma(t_2), \text{ if } t_1 < t_2 \label{eqn:monotone}
\end{align}

The above conditions fix the boundary conditions, and imply that any $\gamma \in \Gamma$ is a monotonically increasing function. This property is also called order-preserving which is important to the current discussion of action recognition as actions are critically dependent on sequencing/ordering of poses/frames. It is easy to show that 
\begin{itemize}
    \item $\forall \gamma_1, \gamma_2 \in \Gamma, \gamma_1 \circ \gamma_2 \in \Gamma,$ 
    \item $\gamma_{Id} \in \Gamma,$  
    \item $\forall \gamma \in \Gamma, \exists \gamma^{-1} \in \Gamma \text{ s.t. } \gamma \circ \gamma^{-1} = \gamma_{Id}$, where $\gamma_{Id}(t) = t$, the identity warping function.
    \end{itemize}
These properties imply that $\Gamma$ admits a group structure with the group action being function composition. We denote by $\dot{\gamma}$, the first derivative of $\gamma \in \Gamma$, or 
\begin{align}
    \gamma(t) = \int_0^t \dot{\gamma}(t) dt, 
    \int_0^1 \dot{\gamma}(t) dt = \gamma(1) - \gamma(0) = 1 \label{eqn:pdf}
\end{align}

Further, due to the monotonically increasing property of $\gamma$, we have $\dot{\gamma} > 0$. This, in conjunction with \eqref{eqn:pdf}, implies that  $\dot{\gamma}$ has the properties of a probability distribution function (positive, and integrates to 1), and the corresponding $\gamma$ is thus equivalent to a cumulative distribution function.

As we work with digitized signals from sensors such as Kinect and mocap, we represent a discrete time series (either signals or features) by $X = \{\mathbf{x}_1, \mathbf{x}_2, \dots, \mathbf{x}_T\}$. In this paper, we will work with time series in $\mathbb{R}^N$. Each $\mathbf{x}_t \in \mathbb{R}^N$ is called a frame of the sequence $X$. More clearly, we have $\alpha(t) = \mathbf{x}_t, t \in \{1,2,\dots, T\}$. The warping function in the case of a discrete time signal is a discretized version of $\gamma \in \Gamma$, which we represent using $\gamma$ with a slight abuse of notation. The derivative $\dot{\gamma}$ can be approximated by first order numerical differencing. Thus,  \eqref{eqn:pdf} now becomes

\begin{align}
    \gamma(t) &= \sum_{i=1}^t\dot{\gamma}(i) \quad \text{ and } \quad  \frac{1}{T}\sum_{i=1}^T \dot{\gamma}(t) = 1. \label{eqn:discretepdf}
\end{align}

Two sequences $\alpha$ and $\beta$ are said to be \textit{equivalent} if there exists a $\gamma \in \Gamma$ such that $\alpha = \beta \circ \gamma$, and the set $\{\alpha \circ \gamma\ | \gamma \in \Gamma\}$ is called the equivalence class of $\alpha$ under rate variations and is denoted by $[\alpha]$. In classical elastic alignment, given two signals, a metric between sequences is defined as the minimal distance between their equivalence classes. This approach can be used to develop class-specific templates, and phase-amplitude separation~\cite{marron2015functional} that reduces intra-class variance, but does not promote inter-class separation. Once the equivalence classes are defined, metrics are designed to compute distances between equivalence classes and develop methods to compute statistical measures such as mean and variance, which can be used to compute optimal alignments~\cite{srivastava2016functional}.

\section{Temporal transformers for learning discriminative warping functions}
\label{sec:temporal_transformer}
\vspace{-0.1in}
The main idea presented in this paper is to use a specialized module, which we call a \textit{Temporal Transformer Network} (TTN) for neural network-based classification which, given an input test sequence $X$, generates a warping function $\gamma$ used to warp the input sequence by computing $X \circ \gamma$ and feed it to the classification network. It is important to note that the warping is carried out using linear interpolation. This makes it possible to train both the TTN and the classifier jointly end-to-end as the entire pipeline is (sub-) differentiable. Another notable aspect of this framework is that the warping functions are predicted without a ``class-template". Even though this sounds paradoxical, we will soon show that this allows our framework to jointly learn features as well as achieve discriminative warps. This capability makes our framework more powerful than template-based matching techniques like Dynamic Time Warping (DTW) and variants~\cite{bellman1958adaptive, sakoe1978dynamic}.

\textbf{Key Insight:} Given two input sequences $X_1$ and $X_2$ such that they differ only by a warping transform, the trained TTN would ideally predict $\gamma_1$ and $\gamma_2$, corresponding to $X_1$ and $X_2$ respectively such that $X_1 \circ \gamma_1 = X_2 \circ \gamma_2$. However, we note that $\gamma_1$ and $\gamma_2$ are not unique because $X_1 \circ \gamma_1 \circ \gamma = X_2 \circ \gamma_2 \circ \gamma, \forall \gamma \in \Gamma$. We believe this to be an important and often overlooked fact, and is formally referred to as invariance to group-action. The non-uniqueness of the warping functions is not fully exploited in the temporal alignment literature. The goal in past works in alignment has been to learn a class-specific template using an iterative optimization method, which is not unique due to invariance to group-action. Non-uniqueness here presents an opportunity that can be exploited to develop discriminative warps for classification problems.

The non-uniqueness of the warping functions can be advantageous as it expands the expressive capacity for classification. Minimizing intra-class variations --- rate variations in our case --- is only one part of the problem. For classification, we would also like to maximize inter-class variations. For example, if we have four sequences $X_1, X_2, X_3$ and $X_4$ such that $X_1, X_2$ belong to Class A, and $X_3, X_4$ belong to class B, the TTN has the capacity to predict $\gamma_1, \gamma_2, \gamma_3$ and $\gamma_4$ such that

\begin{itemize}[leftmargin=*]
    \item $d(X_1 \circ \gamma_1, X_2 \circ \gamma_2) < d(X_1, X_2)$
    \item $d(X_3 \circ \gamma_3, X_4 \circ \gamma_4) < d(X_3, X_4)$
    \item $d(X_i \circ \gamma_i, X_j \circ \gamma_j) >  d(X_i, X_j), i \in \{1,2\} \text{ \& } j \in \{3,4\},$
\end{itemize} 
where $d(.)$ is the Euclidean distance between sequences. However, we do not explicitly train the networks to achieve the above. Both the TTN and the classifier are trained so as to maximize classification performance by minimizing the cross-entropy loss between the predicted and true distribution over the class labels given the input sequence. The TTN can be divided into three sub-modules: 

\begin{figure*}[!t]
    \centering
    \includegraphics[trim={1cm 4cm 2cm 4cm},clip,width=0.95\textwidth]{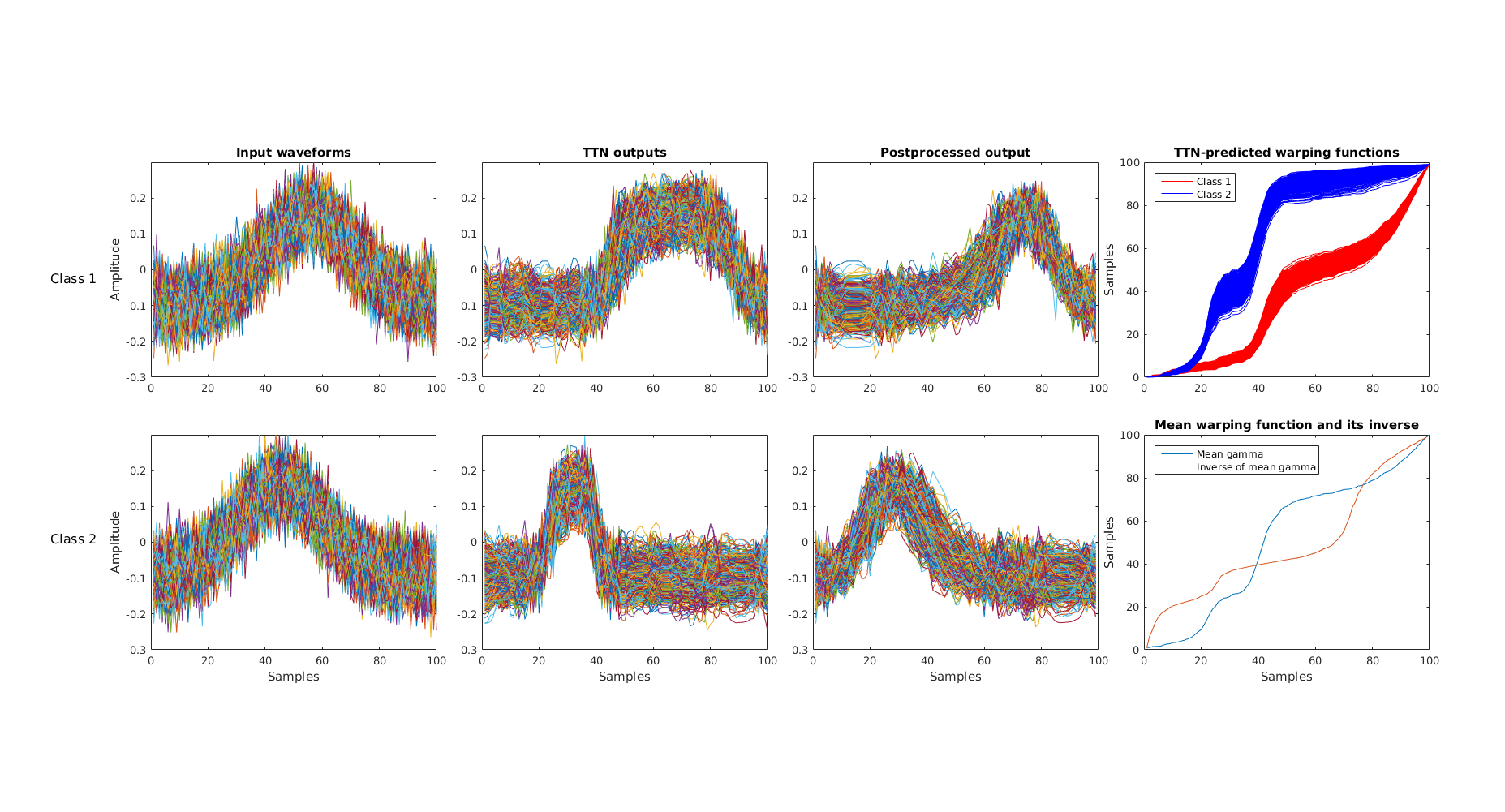}
    \caption{Results on synthetic dataset 1. Rows 1 and 2 show waveforms corresponding to classes 1 and 2 respectively. Columns 1 and 2 show the test inputs and the TTN outputs respectively. It is clear by comparing these columns that the TTN outputs are much better discriminated after warping. The TTN-predicted warping functions also show that the TTN performs class-dependent warping. Column 3 is a better visualization of column 2 after some post-processing making the mean of the generated warping functions $\gamma_\mu$ = $\gamma_{Id}$ (see text).}
    \label{fig:2gaussians}
    \vspace{-0.1in}
\end{figure*}

\textbf{Trainable layers:} As shown in Figure~\ref{fig:ttn_overview}, the input to the TTN trainable layers is an input sequence. The input is then passed through a few layers of convolutions and fully-connected layers. The network outputs a vector of length $T$, such that the first element is set to be zero. $T$ is the length/number of frames in the input sequence. Let us denote this vector by $\mathbf{v} \in \mathbb{R}^{T}$. 

\textbf{Constraint satisfaction layers:} The output $\mathbf{v}$ is unconstrained, and hence, we need to convert it into a valid warping function that satisfies Equation \eqref{eqn:monotone}.

To this end, we first divide $\mathbf{v}$ by its norm to get a unit-vector, followed by squaring each of its entries. This has the effect of converting the vector into a point on the probability simplex. Thus, we use the following mappings:
\begin{equation}
    \dot{\gamma} = \frac{\mathbf{v}}{\| \mathbf{v}\|} \odot \frac{\mathbf{v}}{\| \mathbf{v}\|}, \quad \textrm{and} \quad
    \gamma(t) = T \cdot \sum_{i=1}^t \dot{\gamma}(i), \label{eqn:cumsum}
\end{equation}
where $\odot$ is the Hadamard product (element-wise multiplication). This is treated as the network's estimate of the derivative of the warping function, denoted by $\dot{\gamma}$. We compute the cumulative sum and multiply it by the length of the input sequence, $T$, in order to form the warping function $\gamma$ as shown in Equation \eqref{eqn:cumsum}.

\textbf{Differentiable temporal resampling:} The warping function $\gamma$ is then applied to the input sequence using linear interpolation. We assume that the sampling rate of the signal is high enough in relation to the speed of the activity that simple linear interpolation of the frames is sufficient to get intermediate skeletons to look realistic (in practice, 20 frames/sec is plenty for most common action recognition applications). Warping is performed using the equation $Y(t_t) = X(t_s) = X(\gamma(t_t))$, where $X$ and $Y$ are the input and output sequences respectively, and $t_s$ and $t_t$ are the source and target regular grid indices respectively. The frames of $Y$ are to be defined at regular intervals $t_t = 1,2,\dots,T$. As the $t_s$ corresponding to $t_t$ may not be integers, we use linear interpolation to find the values of $X(\gamma(t_t))$. This operation is sub-differentiable, as in the case of STN. Thus, we can write the expressions of the required gradients as follows (these expressions are adapted from Jaderberg et al.~\cite{jaderberg2015spatial}). If $X^j$ is the input sequence of the $j^{th}$ joint, $X_\tau^j$ is the value at time index $\tau$ for channel $j$, $Y^{j}$ is the warped sequence output by the TTN module and $i \in \{1,2,\dots,T\}$ is the time index, we have
\vspace{-0.1in}

\begin{align}
    \frac{\partial Y_i^j}{\partial X_\tau^j} &= \text{max}(0,1-|t_i^s - \tau|) \text{ and } \\
\frac{\partial Y_i^j}{\partial t_i^s} &= \sum_{\tau=1}^T X^j \cdot
\begin{cases}
    0,& \text{if } |t_i^s - \tau| \geq 1\\
    1,& \text{if } \tau \geq t_i^s\\
    -1,& \text{if } \tau < t_i^s
\end{cases}
\end{align}


\section{Experimental results}
\label{sec:expts}
All networks in this paper are trained and tested using Tensorflow~\cite{abadi2016tensorflow}. Due to space constraints, some training and testing details and results are provided in the supplement.

\begin{figure*}[]
    \centering
    \includegraphics[trim={1cm 4cm 2cm 4cm},clip,width=0.95\textwidth]{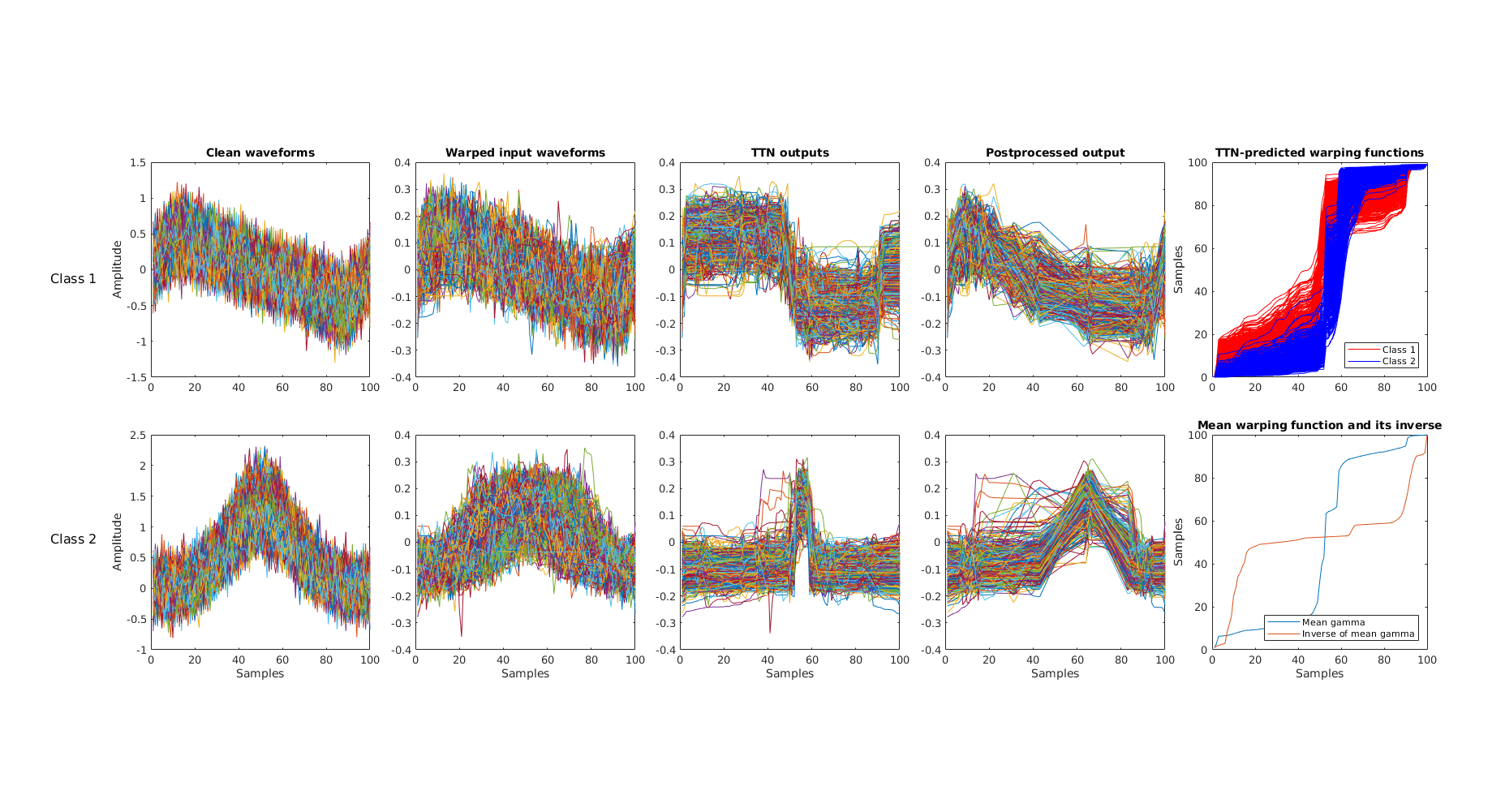}
    \caption{Results on synthetic dataset 2. Rows 1 and 2 show waveforms corresponding to classes 1 and 2 respectively. Columns 1, 2 and 3 show the clean waveforms, test inputs (after random warping) and the TTN outputs respectively. It is clear by comparing these columns that the TTN outputs are much more closely clustered especially for class 2, showing that the TTN outputs are robust to rate-variations. Column 4 is a better visualization of column 3 after some post-processing making the mean of the generated warping functions $\gamma_\mu$ = $\gamma_{Id}$.}
    \label{fig:2classes}
    \vspace{-0.15in}
\end{figure*}

\subsection{Synthetic datasets}

\textbf{(1) Demonstrating discriminative properties of TTN:}
We consider a two-class classification problem where the two classes are one-dimensional time series signals of length 100. Let us denote each sequence in the dataset by $X \in \mathbb{R}^{100}$. All the signals are Gaussian functions with varying amplitude. Signals in class 1 are centered at $t=0.55$ while signals in class 2 are centered at $t=0.45$. Further, we corrupt the function with additive Gaussian noise ($\mathcal{N}(0,0.2)$). Samples of these functions are shown in Figure~\ref{fig:2gaussians}. We generate 8000 training and 2000 test sequences evenly balanced between classes 1 and 2. We use a simple classifier with a one-layer fully connected layer. The TTN is a 2-layer network with 1 convolutional layer producing 1 feature map with a filter of size 8, and 1 fully-connected layer. We train the networks for $10^3$ iterations using Adam optimizer with an initial learning rate of $10^{-4}$ for the classifier. The weights of the TTN are updated at one-tenth the learning rate of the classifier. Figure~\ref{fig:2gaussians} shows the test signals, corresponding outputs of the TTN, as well as the TTN-generated warping functions for every test input. It is clear from the figures that the TTNs predict class-specific warping functions in order to separate the peaks in the signals which makes them more discriminative. Note that this behavior arises automatically by minimizing the cross-entropy loss. In order to visualize the TTN outputs better, we perform post-processing by warping the TTN outputs with $\gamma_{\mu}^{-1}$, where $\gamma_{\mu} = \sum_{i=1}^N \gamma_i$, where $N$ is the size of the test set. This experiment clearly shows that TTNs are effective at increasing inter-class variations, as desired. 

\textbf{(2) Demonstrating rate-invariance of TTN:} Here, we construct a dataset such that rate variations in the signals are the major nuisance parameter. In this scenario, intuitively, minimizing classification error should lead to the following: different signals belonging to the same class, but differing only by a $\gamma$ should come closer to each other after passing through a trained TTN module. In class 1, we have signals which are N-waves with random warping  applied and additive Gaussian noise added to them. Signals in class 2 are similar except that they are Gaussian functions. As before, we generated 8000 training sequences and 2000 test sequences evenly balanced between classes 1 and 2. These are shown before (Column 1) and after random warping (Column 2) in Figure~\ref{fig:2classes}. The TTN, classifier, training and testing protocols are the same as in dataset (1) above. From column 3 in Figure~\ref{fig:2classes}, it is clear the TTN leads to reduction in intra-class rate variations.
\vspace{-0.1in}

\subsection{ICL First-Person Hand Action dataset}
\vspace{-0.1in}
In this section, we conduct experiments on a recently released real-world dataset of hand actions~\cite{FirstPersonAction_CVPR2018}. The dataset contains 3D hand pose sequences with 21 joint locations per frame of 45 daily hand action categories interacting with 26 objects, such as ``pour juice", ``put tea bag" and ``read paper". These sequences are performed by 6 subjects and are acquired using an accurate mocap system. For our experiments, we use the training/test splits suggested by the authors of the dataset~\cite{FirstPersonAction_CVPR2018}, with subjects 1,3,4 used for training and the rest for testing. The training set contains 600 sequences and the test set contains 575 sequences. As the sequences are of varying lengths, we uniformly sample the sequences such that all sequences contain 50 samples. If the sequences are shorter than 50 samples, we use zero-padding. As there are 21 joints per frame, each input sequence is of dimension $50 \times 63$ ($21 \times 3 = 63$). We normalize the sequences such that the wrist position of the first frame is at the origin. We conduct our experiments with two different types of classifiers widely used for action recognition: (1) Temporal Convolutional Network (TCN) and (2) 2-layer LSTM, showing that the proposed TTN framework can yield better results for both the classifier architectures. 

The TTN module consists of 3 fully connected layers with $tanh$ nonlinearity and hidden states of dimensions 16 and 16. The final FC layer produces a vector of length 50 (equal to the input sequence length), with the first element set to zero (see Section \ref{sec:temporal_transformer}).

The TCN architecture contains 1 temporal convolutional layer with 16, 32 or 64 feature maps, and 1 FC layer. We refer to these networks as TCN-16, TCN-32 and TCN-64 respectively. We ran the algorithm 5 times with different initializations of the TTN and report the mean and standard deviation in Table~\ref{table:icl}. 

The LSTM architecture is similar to the one proposed in~\cite{FirstPersonAction_CVPR2018} containing two layers of LSTMs with a state dimension of 1024 and a dropout probability of 0.2. We use momentum optimizer with a momentum of 0.9 for training.

The results obtained are shown in Table~\ref{table:icl}. In addition to our experiments, we have reported results given in~\cite{FirstPersonAction_CVPR2018} for other important algorithms used for 3D pose-based action recognition including JOULE-pose~\cite{hu2015jointly}, Moving Pose~\cite{zanfir2013moving}, Hierarchical Recurrent Neural Networks (HBRNN)~\cite{du2015hierarchical}, Transition Forests (TF)~\cite{garcia2017transition}, and Lie Groups~\cite{vemulapalli2014human} and the Gram Matrix method~\cite{zhang2016efficient}, the last two of which also used DTW for sequence alignment, as well as non-Euclidean features to help improve performance. Among the baseline neural networks, TCN-32 led to the best results for this dataset, and addition of more layers did not yield better performance. We observe that addition of the TTN consistently improves performance over the baseline networks by 3.8 percentage points (TCN-16), 1.0 point (TCN-32), and 2.2 points (TCN-64). In the case of the LSTM classifier, we observe an improvement of 2.25 points using TTN + LSTM over just the LSTM. For the TCN-32 model, we performed K-means clustering (with $\#clusters$ = $\#classes$ and averaged over 100 runs) on the features at the output of the conv. layer of the classifier for the test set, learned with and without TTN. Then, we computed \textit{cluster purity (CP), homogeneity (H) and completeness (C)}. Without the TTN, $CP = 0.519, H = 0.656, C = 0.705$. Adding the TTN module, we get improved scores of $CP = 0.530, H = 0.664, C = 0.709$.

\textbf{Introducing distortions in the data:} 
As datasets are usually collected in lab settings, there are relatively ``clean" and do not contain many rate variations. Now, we introduce artificial rate variations in the data in order to better illustrate the usefulness of the TTN module. Here, we set the sequence length to 100 such that the original sequences of length 50 range from $t = 25$ to $75$, and the rest of the values are set to zero. Now, we apply random ``affine warps" to the training and test data. By an affine warp, we mean a warping function of the form $\gamma(t) = at + b$, $t = 25$ to $75$, which is a linear time-scaling with an offset. We use $a \in [0.75, 1.25]$ and $b \in {0,1,\dots,49}$. 

\begin{figure}
    \centering
    \includegraphics[width=0.4\textwidth]{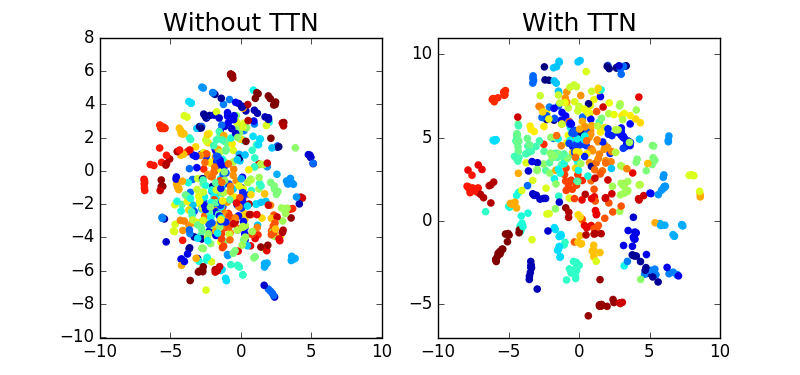}
    \caption{t-SNE plots for the test set features for the affine-warped ICL dataset with and without TTN. When the TTN is employed, we see a better separation of the clusters. This is also reflected in the accuracy scores.}
    \label{fig:tsne}
    \vspace{-0.2in}
\end{figure}

\begin{figure}
    \centering
    \includegraphics[width=0.4\textwidth]{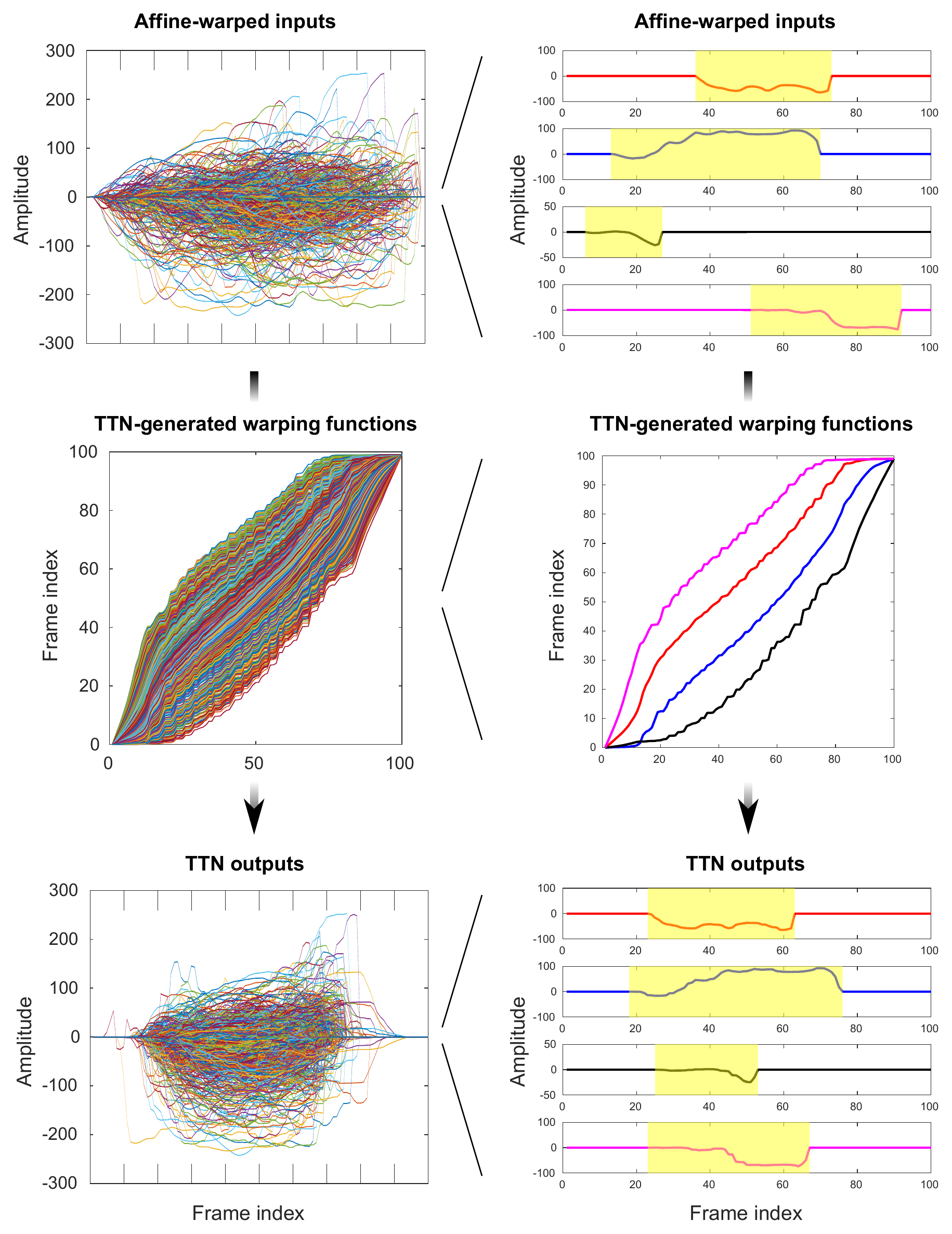}
    \caption{Visualizations of results of TCN-32 + TTN on ICL action dataset with induced rate variations. In the left column are waveforms corresponding to joint 1 of all test sequences. In the right column, 4 of those sequences are shown for clarity. We see clearly that the generated warping functions undo the affine-warp distortion in the test data, and the TTN outputs are nearly perfectly aligned leading to much better classification results.}
    \label{fig:affine_warp}
    \vspace{-0.15in}
\end{figure}

We observe that the induced distortion leads to a huge drop in performance of TCN-32 from 81.74\% to 70.43\%. With the TTN, the performance drop is much lower --- from 82.75\% to 78.26\% and TTN+TCN-32 performs about 8 percentage points higher than TCN-32. Furthermore, from Figure~\ref{fig:affine_warp},  which shows the inputs, generated warping functions and the TTN outputs, it can be readily observed that the TTN performs alignment of the sequences which then makes the classification problem much easier. This experiment shows that addition of the TTN enhances the interpretability of the network, and also delivers superior performance when there are larger rate variations in the data. As before, we performed K-means clustering. Without the TTN, $CP = 0.327, H = 0.487, C = 0.545$. Adding the TTN module, we get improved scores of $CP = 0.476, H = 0.611, C = 0.677$. We also ran t-SNE~\cite{maaten2008visualizing} to visualize the features in 2D. More pure and separated clusters are seen with the addition of TTN, as shown in the Figure~\ref{fig:tsne}.

\begin{table}[]
    \centering
    \begin{tabular}{c c}
        \hline
         Method  & Accuracy (\%)\\
         \hline
         Moving Pose~\cite{zanfir2013moving} & 56.34 \\
         JOULE-pose~\cite{hu2015jointly}& 74.60 \\
         HBRNN~\cite{du2015hierarchical} & 77.40\\
         TF~\cite{garcia2017transition} & 80.69\\
         Lie Group~\cite{vemulapalli2014human} & 82.69\\
         Gram Matrix~\cite{zhang2016efficient} & \textit{\textbf{85.39}}\\
         \hline
         2-layer LSTM & 76.17\\
         2-layer LSTM + TTN & \textbf{78.43}\\
         \hline
         TCN-16 & 76.28 $\pm$ 0.29\\
         TCN-16 + TTN & \textbf{80.14 $\pm$ 0.33}\\
         \hline
         TCN-64 & 79.10 $\pm$ 0.76\\
         TCN-64 + TTN & \textbf{81.32 $\pm$ 0.36}\\
        \hline
         TCN-32 & 81.74 $\pm$ 0.27\\
         TCN-32 + TTN & \textbf{82.75 $\pm$ 0.31}\\
        \hline
         TCN-32 (affine warp) & 70.43 \\
         TCN-32 + TTN (affine warp) & \textbf{78.26}\\
         \hline
        
    \end{tabular}
    \caption{Action recognition results on the ICL hand action dataset showing that LSTM+TTN and TCN+TTN frameworks consistently outperform LSTM and TCN baselines.}
    \label{table:icl}
    \vspace{-0.1in}
\end{table}

\subsection{NTU RGB-D dataset}
\vspace{-0.1in}
In this section, we conduct experiments on a large-scale dataset of human actions known as the NTU RGB-D dataset~\cite{Shahroudy_2016_CVPR} which contains about 56000 sequences of 3D skeleton positions acquired by a Microsoft Kinect. 25 joint locations are provided for each skeleton. The dataset contains actions belonging to 60 human activities performed by 45 subjects, with some actions containing two actors. The data are acquired using a Microsoft Kinect. We sample 50 frames per sequence uniformly. We conduct two sets of experiments for this dataset -- Cross Subject (CS) and Cross View (CV) -- as per the protocol suggested by the authors in~\cite{Shahroudy_2016_CVPR} using the same training and testing splits. 

We construct a TTN module with 2 temporal convolutional layers and 3 FC layers with ReLU non-linearity. We use a filter size of 8 and 16 output feature maps in each conv layer. The FC layers produce hidden representations of sizes 16, 16 and 50 respectively.

We use the Temporal Convolution Network (TCN) described in~\cite{kim2017interpretable}. The network consists of 10 convolutional layers with batch normalization and ReLU non-linearity. While training, the TTN parameters are updated at one-tenth the learning rate of the TCN.

The results obtained for this dataset are shown in Table~\ref{table:ntu}. For cross-subject experiments, we observe that the addition of the TTN module results in a performance improvement of about 1 percentage point over the baseline TCN. We also found that using 2 parallel TTNs and concatenating the TTN outputs results in further improvement with a final performance of $77.80 \%$. The addition of the TTN module leads to less improvement in the case of cross-view experiment. This can be explained by the fact that there is likely less rate variation in the case of cross-view protocol compared to cross-subject. The ablation studies for TTN and TCN are provided in the supplement. As is the case of the ICL dataset, we performed K-means clustering (with $\#clusters$ = $\#classes$ and averaged over 100 runs) on the features obtained at the penultimate layer of the TCN, and computed the same clustering metrics. Without the TTN, $CP = 0.466, H = 0.575, C = 0.597$. Adding the TTN module, we get improved scores of $CP = 0.493, H = 0.596, C = 0.621$. The corresponding t-SNE plots are provided in the supplement showing improved clusters with the addition of the TTN module.
\vspace{-0.1in}

\begin{table}[]
    \centering
    \begin{tabular}{c c c}
        \hline
         Method  & CS (\%) & CV (\%)\\
         \hline
         Lie Groups~\cite{vemulapalli2014human} & 50.08 & 52.76 \\
         FTP Dynamic Skeletons~\cite{hu2015jointly} & 60.23 & 65.22 \\
         HBRNN~\cite{du2015hierarchical} & 59.07 & 63.97 \\
         2-layer part-LSTM~\cite{Shahroudy_2016_CVPR} & 62.93 & 70.27 \\
         STA-LSTM~\cite{song2017end} & 73.40 & 81.20\\
         VA-LSTM~\cite{zhang2017view} & 79.40 & 87.60 \\
         STA-GCN~\cite{yan2018spatial} & \textit{\textbf{81.50}} & \textit{\textbf{88.30}} \\
         \hline
         TCN~\cite{kim2017interpretable} & 76.54 & 83.98 \\
         TCN + TTN & \textbf{77.55} & \textbf{84.25} \\
         \hline
        
    \end{tabular}
    \caption{Action recognition results on the NTU RGB-D dataset showing that TCN+TTN frameworks outperforms the TCN.}
    \label{table:ntu}
   \vspace{-0.2in}
\end{table}

\section{Discussion and future work}
\vspace{-0.05in}
In this work, we have proposed the Temporal Transformer Network (TTN) which can be readily integrated into classification pipelines. TTN has the ability to generate rate-invariant as well as discriminative warping functions for general time-series classification. We have shown improved classification results using different types of classifiers -- TCNs and LSTMs -- on challenging 3D action recognition datasets acquired using different modalities -- Kinect and mocap. We have demonstrated the rate-invariant and discriminative properties of the TTN. 

In the future, we would like to apply the ideas presented in this paper to video action recognition. However, it is not immediately clear how to perform temporal warping for videos as the currently widely used features for video frames may not be well suited for interpolation. One possible solution is to jointly train the image-level features and the action classification pipeline along with the TTN module. Temporal transformers can also be applied in general time-series classification which includes recognition from wearables, speech, EEG data, etc. Unsupervised pattern discovery with inbuilt warp-invariant metrics will be another interesting direction for further research.

\vspace{-0.1in}

\noindent \paragraph*{Acknowledgements:} This work was supported in part by NSF grant 1617999, and ARO grant number W911NF-17-1-0293

{\small
\bibliographystyle{ieee}
\bibliography{egbib}
}


\clearpage

\textbf{\Large{Supplementary Material}}

\section{Additional experimental results and details}

\subsection{Additional synthetic dataset Demonstrating rate-invariance property of TTN:}
In this case, we construct a dataset such that rate variations in the signals are the major nuisance parameter. In this scenario, intuitively, minimizing classification error should lead to the following: different signals belonging to the same class, but differing (approximately) only by a $\gamma$ should come closer to each other after passing through a trained TTN module. In class 1, we have signals which are a a mixture of two Gaussian functions with random warping  applied and additive Gaussian noise added to them. Signals in class 2 are similar except that they are a single Gaussian function with the same mean and variance. As before, we generated 8000 training sequences and 2000 test sequences evenly balanced between classes 1 and 2. These are shown before (Column 1) and after random warping (Column 2) in Figure~\ref{fig:2classes}. The TTN, classifier and the training and testing protocol are the same as in dataset (1) above. From column 3 in Figure~\ref{fig:2classes}, it is clear the TTN leads to reduction in intra-class rate variations. Table~\ref{table:2classes} shows the classification accuracies obtained with and without the TTN module (averaged over 10 runs). When no warping is present in the input data, both variants yield perfect accuracy. When warping is introduced in the dataset, the performance of the vanilla model (i.e., without TTN) drops significantly. With the addition of the TTN module, most of the lost performance can be recovered.

\begin{figure*}[]
	\centering
	\includegraphics[trim={1cm 4cm 2cm 4cm},clip,width=\textwidth]{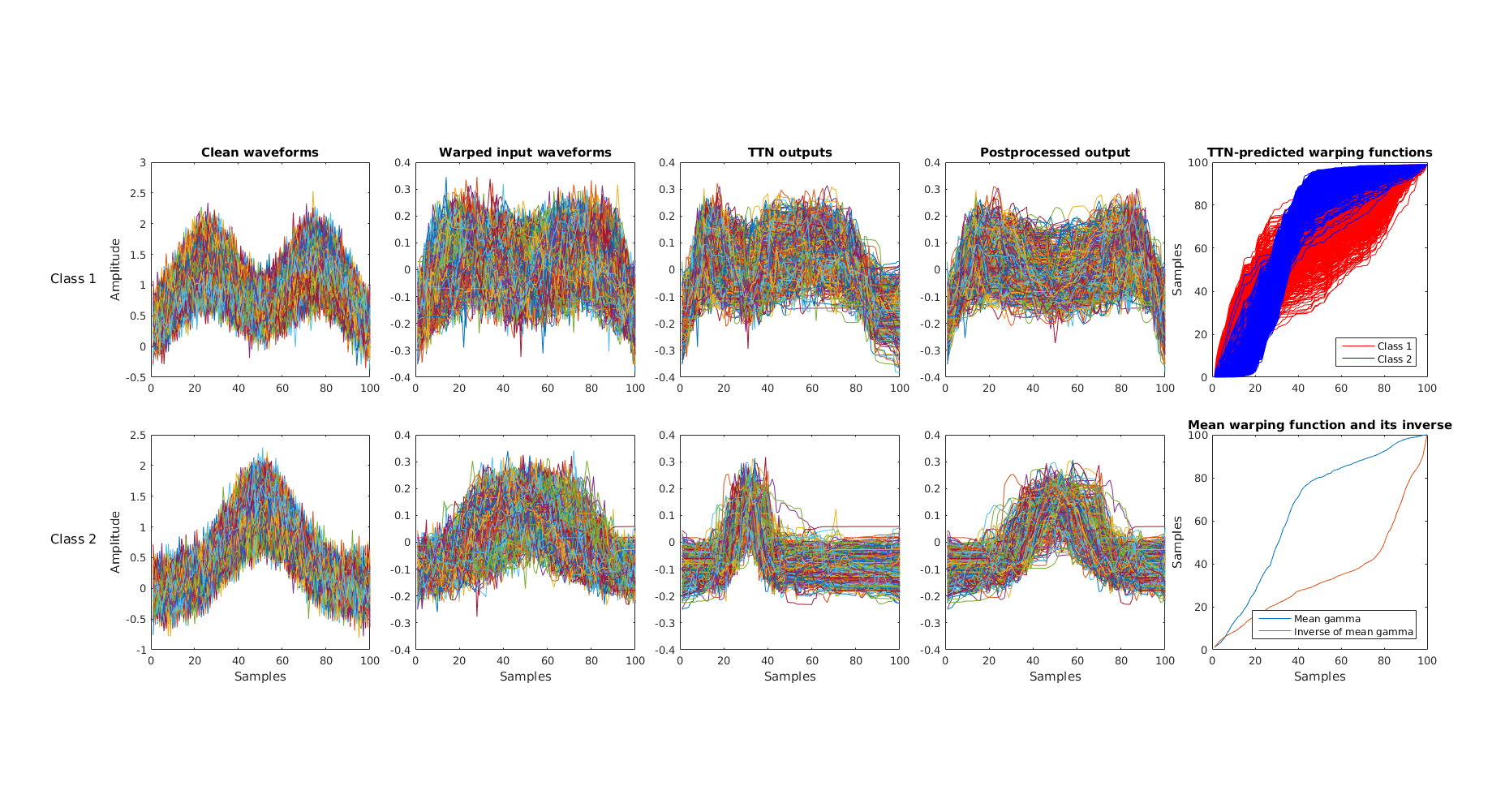}
	
	\caption{Results on synthetic dataset 2. Rows 1 and 2 show waveforms corresponding to classes 1 and 2 respectively. Columns 1, 2 and 3 show the clean waveforms, test inputs (after random warping) and the TTN outputs respectively. It is clear by comparing these columns that the TTN outputs are much more closely clustered especially for class 2, showing that the TTN outputs are robust to rate-variations. Column 4 is a better visualization of column 3 after some post-processing making the mean of the generated warping functions $\gamma_\mu$ = $\gamma_{Id}$.}
	\label{fig:2classes}
\end{figure*}

\begin{table}[h]
	\centering
	\begin{tabular}{c c c}
		\hline
		& Vanilla  & TTN\\
		\hline
		Unwarped & 100.00 $\pm$ 0.00 \% & \textbf{100.00 $\pm$ 0.00} \%\\
		\hline
		Warped & 96.31 $\pm$ 0.021 \% & \textbf{99.03 $\pm$ 0.15} \% \\
		\hline
		
	\end{tabular}
	\caption{Recognition results (\%) for synthetic dataset 2. Addition of TTN clearly outperforms the baseline.}
	\label{table:2classes}
	\vspace{-0.2in}
\end{table}

\subsection{ICL First-Person Hand Action Dataset \cite{FirstPersonAction_CVPR2018}}
\textbf{Training protocol details omitted in the main paper in Section 5.2:}
Both the classifier and the TTN are trained with momentum optimizer with momentum set to 0.9. A batch size of 16 is used and the networks are trained for 50000 iterations. We use an initial learning rate of $10^{-3}$ for the classifier and and $10^{-4}$ for the TTN. The learning rate is reduced to one-tenth after 35000 and 45000 iterations. 

\subsection{NTU RGB-D Dataset}
\textbf{TCN architecture and training details}: We use the Temporal Convolution Network (TCN) described in \cite{kim2017interpretable}. The network consists of 10 convolutional layers with batch normalization and ReLU non-linearity. The network can be divided into 3 blocks of conv layers. After each block, the first conv layer of the next block is of stride 2 such that the inputs for successive block are of length half of that of the previous block. Residual connections are employed in the network as well. All the convolutional filters are of size 8. The convolutional layers produce 64,128 and 256 feature maps for blocks 1, 2 and 3 respectively. We use momentum optimizer with momentum 0.9 and initial learning rate of $5 \times 10^{-3}$. The network is trained for $2\times 10^{5}$ iterations with a batch size of 72, and the learning rate was reduced one-tenth the value after $10^5$ and $1.5 \times 10^5$ iterations. We found this to be these values of the hyperparameters to be the optimal setting for the baseline classifier network. While training, the TTN parameters are updated at one-tenth the learning rate of the TCN.

\textbf{Ablation study for the TCN:}
We study the effect of the number of layers in the classifier network on the performance. The TCN architecture consists of 3 blocks of conv layers. We remove 1 block at a time and compare the cross-subject classification rate.
The results are shown in Table~\ref{table:ntu_tcn_ablation}. We see that the addition of the TTN produces better results in all cases.

\begin{table}[]
	\centering
	\begin{tabular}{c c c}
		\hline
		Method  & w/o TTN (\%) & w/ TTN (\%)\\
		\hline
		TCN (4 layers) & 70.72 & \textbf{71.63} \\
		TCN (7 layers) & 75.06 & \textbf{75.30} \\
		TCN (10 layers) & 76.54 & \textbf{77.55} \\
		\hline
		
	\end{tabular}
	\caption{Ablation results on the TCN for the NTU database. Cross-subject action recognition results show that the TTN+TCN consistently performs better than TCN for different sizes of TCN.}
	\label{table:ntu_tcn_ablation}
\end{table}

\textbf{Ablation study for the TTN module:}
We study the effect of the number and type (convolutional or fully-connected) of layers in the TTN on the cross-subject classification performance.
The results are shown in Table \ref{table:ntu_ttn_ablation}. We see that the architecture of the TTN indeed has a significant effect on the performance. However, irrespective of the TTN architecture, addition of the TTN produces better performance in all cases.

\begin{figure}
	\centering
	\includegraphics[width=0.45\textwidth]{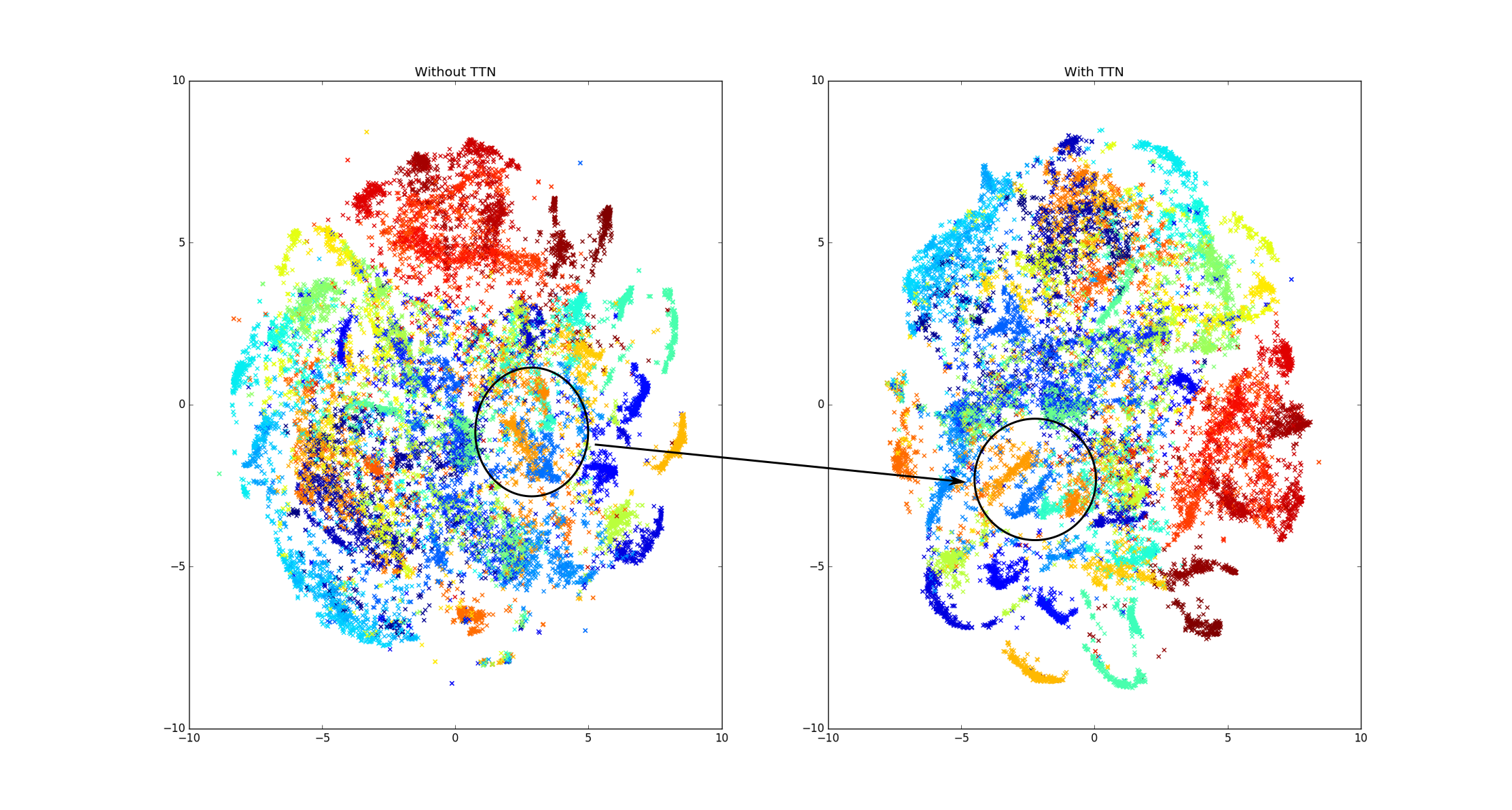}
	\caption{t-SNE plots for the test set features for the NTU dataset with and without TTN. When the TTN is employed, we see better separation in the clusters.}
	\label{fig:tsne_ntu}
\end{figure}

\textbf{t-SNE plot:} Using the features of the last layer of the TCN with 10 layers, we have computed the t-SNE plots with and without TTN for the NTU cross-subject test set. When the TTN is employed, we see better separation in the clusters as shown in Figure~\ref{fig:tsne_ntu}.

\begin{table}[]
	\centering
	\begin{tabular}{c c}
		\hline
		Method  & Accuracy (\%)\\
		\hline
		Baseline TCN \cite{kim2017interpretable} & 76.54 \\
		\hline
		2 conv + 3 FC &  \textbf{77.55}\\
		1 conv + 3 FC &  77.26\\
		0 conv + 3 FC &  76.90\\
		2 conv + 2 FC &  77.20\\
		2 conv + 1 FC &  77.12\\
		\hline
		
	\end{tabular}
	\caption{Ablation results on the TTN for the NTU database using TCN as the classifier \cite{kim2017interpretable}. Cross-subject action recognition results show that the TTN+TCN consistently performs better than TCN for different sizes of TTN.}
	\label{table:ntu_ttn_ablation}
\end{table}

\end{document}